\documentclass[lettersize,journal]{IEEEtran}
\usepackage{amsmath,amsfonts}
\usepackage{algorithmic}
\usepackage{algorithm}
\usepackage{array}
\usepackage[caption=false,font=normalsize,labelfont=sf,textfont=sf]{subfig}
\usepackage{textcomp}
\usepackage{stfloats}
\usepackage{url}
\usepackage{verbatim}
\usepackage{graphicx}
\usepackage{cite}
\usepackage[colorlinks,linkcolor=blue]{hyperref}

\usepackage{multirow}
\usepackage{array, makecell, caption}
\usepackage{times}
\usepackage{epsfig}
\usepackage{graphicx}
\usepackage{amsmath}
\usepackage{amssymb}
\usepackage{pifont}
\usepackage{epstopdf}
\usepackage{booktabs}

\newcommand{\customref}[2]{%
  \hyperref[#1]{\textbf{#2}}%
}
\usepackage{amssymb}
\usepackage{pifont}
\usepackage{fontawesome}
\usepackage[table]{xcolor}
\definecolor{color0}{RGB}{211,211,211} 
\definecolor{color1}{RGB}{175,238,238} 
\definecolor{color2}{RGB}{225,255,255} 
\definecolor{color3}{RGB}{255,215,0} 
\definecolor{color4}{RGB}{255,248,220} 
\definecolor{color5}{RGB}{255,160,122} 
\definecolor{color6}{RGB}{255,228,225} 
\usepackage{multirow}
\usepackage{color}
\usepackage{hyperref}
\hypersetup{
    colorlinks=true,
    linkcolor=blue,
    filecolor=blue,      
    urlcolor=blue,
    citecolor=cyan,
}

\begin{document}

\title{GEM: Boost Simple Network for Glass Surface Segmentation via Vision Foundation Models}


\author{Jing Hao, Moyun Liu,~\IEEEmembership{Graduate Student Member,~IEEE}, Jinrong Yang, and Kuo Feng Hung

\thanks{\emph{(Corresponding author: Jing Hao)}}
\thanks{\emph{(Jing Hao and Moyun Liu contributed to this paper equally.)}}

\thanks{Jing Hao, Moyun Liu, and Jinrong Yang are with the School of Mechanical Science and Engineering,
Huazhong University of Science and Technology, Wuhan 430074, China
(e-mail: isjinghao@gmail.com, lmomoy@hust.edu.cn, and yancieyjr@gmail.com).}
\thanks{Kuo Feng Hung is with the University of Hongkong, Hongkong, China (e-mail: hungkfg@hku.hk).}
}







\maketitle

\begin{abstract}
Detecting glass regions is a challenging task due to the inherent ambiguity in their transparency and reflective characteristics. 
Current solutions in this field remain rooted in conventional deep learning paradigms, requiring the construction of annotated datasets and the design of network architectures. 
However, the evident drawback with these mainstream solutions lies in the time-consuming and labor-intensive process of curating datasets, alongside the increasing complexity of model structures.
In this paper, we propose to address these issues by fully harnessing the capabilities of two existing vision foundation models (VFMs): Stable Diffusion and Segment Anything Model (SAM).
Firstly, we construct a Synthetic but photorealistic large-scale Glass Surface Detection dataset, dubbed S-GSD, without any labour cost via Stable Diffusion. 
This dataset consists of four different scales, consisting of 168k images totally with precise masks. 
Besides, based on the powerful segmentation ability of SAM, we devise a simple \textbf{G}lass surface s\textbf{E}g\textbf{M}entor named GEM, which follows the simple query-based encoder-decoder architecture. 
Comprehensive experiments are conducted on the large-scale glass segmentation dataset GSD-S. 
Our GEM establishes a new state-of-the-art performance with the help of these two VFMs, surpassing the best-reported method GlassSemNet with an IoU improvement of 2.1\%. 
Additionally, extensive experiments demonstrate that our synthetic dataset S-GSD exhibit remarkable performance in zero-shot and transfer learning settings.
Codes, datasets and models are publicly available at:
\href{https://github.com/isbrycee/GEM/}{https://github.com/isbrycee/GEM.}
\end{abstract}


\begin{figure}
\centerline{\includegraphics[width=0.48\textwidth]{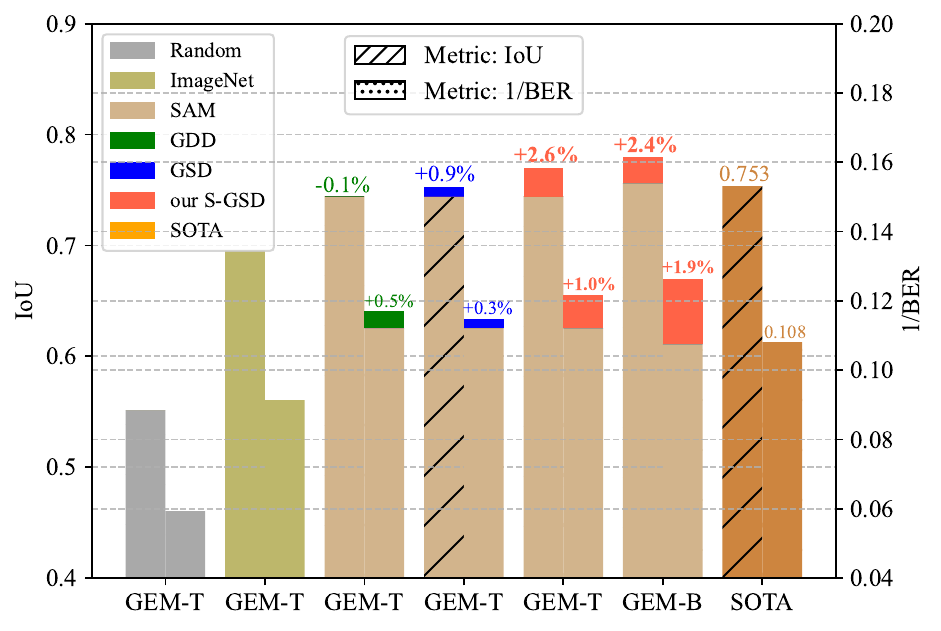}}
\caption{The comparison of the metrics IoU (Intersection over Union) and 1/BER (balance error rate) on GSD-S validation set upon different pre-trained datasets and the state-of-the-art (SOTA).  Our S-GSD dataset boosts higher improvements compared with two real glass surface datasets, GDD and GSD. The GEM-T and GEM-B indicate the GEM-Tiny and GEM-Base, respectively. The SOTA is derived from GlassSegNet \protect\cite{lin2022exploiting}. The metric IoU represents the accuracy of localization, and the metric BER refers to the ratio of errors in classification.}
\label{fig:iou_ber}
\end{figure}

\section{Introduction}

\IEEEPARstart{G}{lass} surfaces, including glass doors, windows, and walls of modern architecture, are becoming fashionable for aesthetic appeal and energy-efficient interior lighting. These glass-made items seem unfriendly towards those intelligent robots or unmanned planes operating automatically because their autonomous systems typically lack the ability to identify ubiquitous but transparent glass surfaces. The glass region nearly has no fixed patterns; that is, the transparent glass share the same colors and appearances with arbitrary objects from the background, leading inconsistent visual features if the background or viewpoint is changed. This inherent property of transparency poses challenges for visual recognition methods. 

With the advent of deep learning techniques, numerous researchers have propelled the development of glass segmentation from both the aspects of data and model architecture design.
From the data perspective, it is widely acknowledged that the quantity and quality of data utilized for training neural networks play a crucial role in determining the performance of the models \cite{meng2020coupled}. 
Currently, there exist three mainstream open-source glass surface segmentation datasets, including GDD \cite{mei2020don}, GSD \cite{lin2021rich}, and GSD-S \cite{lin2022exploiting}. 
However, these datasets are characterized by their limited scale (approximately 4k samples) and are challenging to scale up due to the time-consuming and labor-intensive nature of the image sampling and manual annotation processes. 

From the model architecture perspective, some efforts have been made to enrich the representation of glass by incorporating contextual information \cite{lin2022exploiting}, reflection detection \cite{lin2021rich}, and boundary supervision \cite{he2021enhanced}. Furthermore, in pursuit of more robust glass area segmentation, several methods aim to exploit alternative cues such as depth \cite{wang2013glass}, light-field \cite{xu2015transcut}, polarized light \cite{kalra2020deep}, intensity \cite{mei2022glass}, and thermal imaging \cite{huo2023glass}. However, these approaches exhibit two notable drawbacks. Firstly, increasingly complex model architectures inevitably results in larger parameter sizes and computational complexities, consequently decreasing model inference speeds. Secondly, obtaining these additional modal data in practical application scenarios is not straightforward, thereby escalating the deployment costs of the system.

Recently, with the emergence of various Visual Foundation Large Models (VLMs), AI research has witnessed a paradigm shift. The VLMs, such as the Segment Anything Model (SAM) \cite{kirillov2023segment}, the Depth Anything Model \cite{yang2024depth}, GPT-4v \cite{yang2023gpt4v}, and the Stable Diffusion model \cite{rombach2022high}, have shown particular promise in specific domains. Some researchers have built upon these VFMs to address key challenges across multiple specific task scenarios. VFMM3D \cite{ding2024vfmm3d} achieved high-precision monocular 3D object detection by integrating SAM and the Depth Anything Model. FoundationGrasp \cite{tang2024foundationgrasp} realized task-oriented grasping tasks by harnessing the open-ended knowledge from GPT-4 \cite{achiam2023gpt4} and CLIP \cite{radford2021clip}. These works validate that the flexible and appropriate exploration of potential inherent in foundational models can overcome the challenges and limitations of traditional deep learning solutions. Inspired by this new AI research paradigm, we embarked on an initial exploration into harnessing VFMs for the glass segmentation task. Leveraging the prior knowledge inherent in VFMs, we successfully addressed challenges associated with the acquisition of data and the complexity of model structures for segmenting glass regions. In addition, we achieved a new state-of-the-art performance on the public glass segmentation dataset GSD-S, as shown in Fig. \ref{fig:iou_ber}. 

Initially, we synthesize a large-scale and photorealistic glass segmentation dataset named S-GSD using the prior knowledge of Stable Diffusion and the controllability of ControlNet, mitigating the problem of data scarcity to some extent. The Stable Diffusion is trained at web-scale images, and it incorporates the text-based guidance with the image generation process. The ControlNet \cite{zhang2023controlnet} adds spatially localized input conditions to Stable Diffusion model, enabling the statistical distribution of generated image to be more consistent with localized conditions. We firstly finetune the ControlNet on a public glass segmentation dataset so as to transfer the general knowledge of ControlNet into the glass scenario distribution, and then synthesize numerous images by using the precise mask prior in the public dataset as a control condition. 
Our synthetic dataset comprises four distinct scales, including 4k, 24k, 47k, and 94k image-mask pairs, respectively, with no overlap among the datasets at each scale.
This dataset encompasses a diverse array of contents and conforms to the distribution observed in real images, and it is used in the manner of zero-shot and transfer learning. 
Additionally, we transferred the precise segmentation capability of SAM to delineate the foreground glass regions, employing a query-based encoder-decoder architecture to decode the features extracted by SAM.
By amalgamating the SAM with a simple mask decoder through some simple modifications, we design a simple \textbf{G}lass surface s\textbf{E}g\textbf{M}entor, dubbed GEM. 
Specifically, our GEM only consists of a ViT \cite{dosovitskiy2020image} backbone, a simple feature pyramid \cite{li2022vitdet}, a discerning query selection module, and a mask decoder. 
We explore two different scales of feature extractors, the image encoder of SAM and MobileSAM \cite{zhang2023faster}, as the backbone for GEM.
The simple feature pyramid aims to construct multi-scaled features tailored for the dense prediction tasks from a single-scale feature map upon ViT architecture. 
To further address the difficulty arising from the similarity between glass and its surrounding environment, we also devise a discerning query selection module. By pre-predicting the output mask through the multi-scale features, those features with high confidence scores can be used to initialize the query in the mask decoder for further refinement. 
The framework of GEM is streamlined because not only it does not need the contextual relationships of scenes or object boundary prior as inputs but it does not require extra cues like depth, polarized light, or the thermal image as inputs.

\begin{figure*}[!t]
\setlength{\abovecaptionskip}{-15 pt}
\begin{center}
\centerline{\includegraphics[width=18cm]{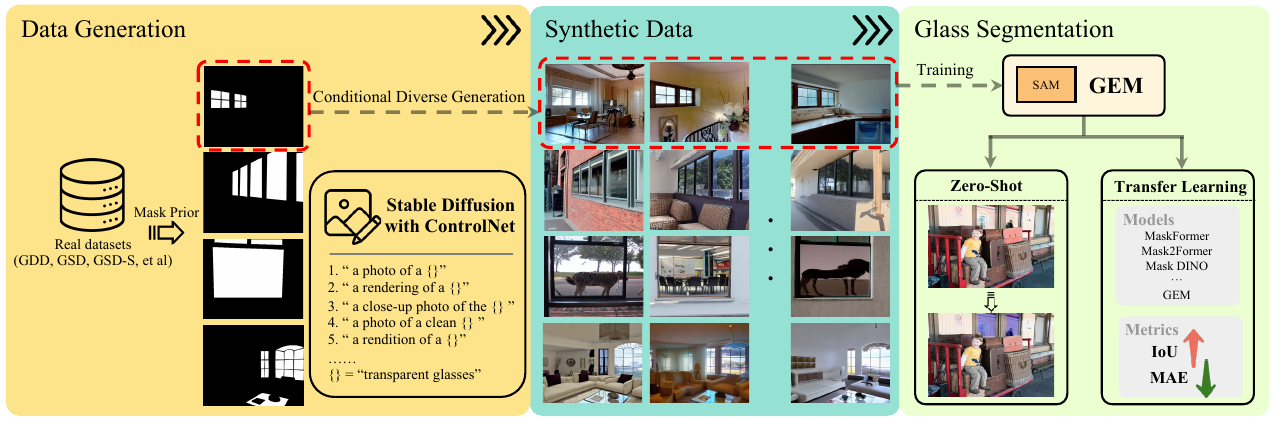}}
\end{center}
\caption{The pipeline of glass surface segmentation with the help of VFMs. Firstly, we utilize the ControlNet with Stable Diffusion to generate massive high-quality images by using the mask prior in the real dataset as control conditions. After that, we can get large-scale and diverse image-mask paris. Finally, we train proposed GEM model on the synthetic dataset and implement zero-shot and transfer learning. }
\label{fig:pipeline}
\end{figure*}

The whole pipeline of harnessing VFMs for glass surface segmentation is demonstrated in Fig. \ref{fig:pipeline}. We evaluate our GEM on the public glass segmentation dataset GSD-S, and it surpasses the previous state-of-the-art by a large margin (IoU +2.1\%). Moreover, compared with two public glass surface datasets, GDD and GSD, pretraining on S-GSD shows excellent zero-shot \& finetuning performance on GSD-S, which demonstrates the high-quality and reliability of S-GSD auto-generated by VFMs. Furthermore, after utilizing the pre-trained models pretrained on synthetic S-GSD, the metric IoU of GEM-Tiny and GEM-Base can be improved by 2.6\% and 1.8\%, respectively. 
To summarize, our contributions are fourfold:
\begin{itemize}
    \item Our method is the first work that integrates VFMs to enhance the glass surface segmentation. By utilizing the powerful abilities of Stable Diffusion and SAM, we address the challenges inherent in traditional deep learning solutions, including data collection, manual annotation, and complex structural design.
    \item We automatically construct and release a large-scale glass segmentation dataset S-GSD, comprising 168k images with precise masks. This dataset is entirely synthetic, without any manual annotation process, yet it exhibits high levels of reliability, rich diversity, and mask quality.
    \item Our proposed GEM transfers the precise segmentation capability of SAM to delineate the glass regions by employing a query-based encoder-decoder architecture. It achieves higher precision on glass segmentation with fewer model parameters and FLOPs (floating point operations).
    \item Extensive experiments show that our GEM surpasses the previous state-of-the-art methods within a smaller model size and faster inference ability (IoU +1.7\%, 2/3 Params, 1/12 FLOPs, 2.9× faster FPS).
    Also, our S-GSD exhibits substantial improvement on zero-shot learning and transfer learning settings.
\end{itemize}


\section{Related works}
\label{Related works}
\noindent
\subsection{Glass Surface Segmentation}
Due to the ubiquity of glass in daily scenes, accurate glass segmentation is vital for the safety of autonomous systems. However, this task is challenged by a shortage of annotated data. Although datasets such as those introduced by Mei et al. \cite{mei2020don}, Lin et al. \cite{lin2022exploiting}, and Lin et al. \cite{lin2021rich} provide real-world examples, their sizes remain notably limited. Consequently, researchers have turned to various strategies to improve the performance of segmentation models. One approach involves leveraging additional cues such as reflection \cite{lin2021rich}, intensity \cite{mei2022glass}, polarization \cite{mei2022glass}, depth \cite{mei2021depth}, or thermal properties \cite{huo2023glass} to enrich the representation of glass surfaces. However, these models often exhibit limited generalizability as they heavily rely on these supplementary features. To address this limitation and develop more robust models, some studies have focused on designing high-complexity architectures. For instance, Lin et al. \cite{lin2022exploiting} introduce lots of attention modules to capture semantic relations within the environment. Given the inherent transparency of glass, distinguishing glass regions becomes challenging. Therefore, Han et al. \cite{han2023internal} emphasize improved boundary perception in their work. Similarly, Fan et al. \cite{fan2023rfenet} incorporate complex structures that integrate semantic and boundary features, with a focus on refining ambiguous points to achieve more precise glass surface segmentation.

\noindent
\subsection{Vision Foundation Models}
In recent years, significant progress has been made in the development of foundational vision models. CLIP \cite{radford2021clip} uses image-text data to obtain a pre-training foundational model, and it achieves great zero-shot performance for visual tasks. To enhance the efficiency and stability of CLIP, masked contrastive learning is utilized in FLIP \cite{li2023scaling} and EVA \cite{fang2023eva}. To make image perception more interactive, many conversational vision-language models \cite{achiam2023gpt} \cite{zhu2023minigpt} are proposed to input both image and text, and they are capable to reason and understand the holistic image. Benefit from large-scale pre-training, image segmentation model achieves superior zero-shot performance \cite{wang2023seggpt} \cite{zou2024segment}. SAM \cite{kirillov2023segment} uses an image encoder to generate image embeddings. In the meantime, a prompt encoder and a mask decoder are employed to produce high-quality masks. SAM extremely helps downstream visual tasks based on its powerful representation ability. To achieve high-quality image generation, Stable Diffusion \cite{rombach2022high} introduces latent space to replace pixel space. DALL-E \cite{ramesh2021zero} achieves great generalization with a scale autoregressive transformer for zero-shot text-to-image generation. Besides, DALL-E2 \cite{ramesh2022hierarchical} extends the latent space for both image and text, and the text-image latent prior is proven to make a huge contribution to the generation. To support additional input conditions for large pretrained diffusion models, ControlNet \cite{zhang2023adding} was proposed to augment models like Stable Diffusion to enable conditional inputs.


\noindent
\subsection{The Applications of Vision Foundation Models}
Due to the powerful representation of VFMs, many works try to introduce them in their tasks. For example, SAM has been commonly used in medical image \cite{he2023accuracy}, remote sensing \cite{osco2023segment}, and defect detection \cite{ahmadi2023application}. Recently, the SAM \cite{kirillov2023segment} is also evaluated in the glass detection field \cite{han2023segment}, and it proves the visual foundation model cannot directly segment glass without any modification. In the meantime, with the advent of AIGC, generative models have been exploited to synthesize data for self-supervised pre-training and few/zero-shot learning, highlighting the transfer learning capacity of synthetic training data. He et al. \cite{he2022synthetic} have pointed out that the synthetic data delivers superior transfer learning performance for large-scale model pre-training, even outperforming ImageNet pre-training. Sariyildiz et al. \cite{sariyildiz2023fake} shows that models trained on synthetic images exhibit strong generalization properties and perform on par with models trained on real data for transfer. Marathe et al. \cite{marathe2023wedge} establishes a synthetic dataset named WEDGE that can be used to finetune state-of-the-art detectors, improving SOTA performance on real-world benchmarks. 


\section{Proposed Synthetic Dataset}
\label{Methodology}

\begin{table*}[]
\centering
\caption{The detailed comparison of our proposed S-GSD dataset, compared with three real glass datasets, in terms of image source, mask source, image number, and image complexity. ``1x'' means the number of synthetic image is 1 times to the real dataset GSD-S \cite{lin2022exploiting}. The image complexity \cite{feng2022ic9600} can be considered as the amount of detail and variety in the image, where a higher numerical value indicates a greater degree of data complexity. Note that there is no overlap between the four datasets at different scales.
}
\begin{tabular}{l|c|c|c|cccc}
\toprule
Dataset & Image Source & Mask Source & Number &  Image Complexity \cite{feng2022ic9600} \\
\midrule
GDD     & Manually Captured & Manually Annotated & 3916 & 0.4015 \\
GSD     &  Public datasets \& Manually Captured & Public datasets \& Manually Annotated & 4102 & 0.4948 \\
GSD-S   & Public datasets & Manually Relabled & 4519 & 0.5052 \\
\midrule
Our S-SGD-1x & \multirow{4}{*}{Auto-generated by Stable Diffusion with ControlNet} & \multirow{4}{*}{Mask prior in GSD-S Training Set} & 3912 & \multirow{4}{*}{0.4317} \\
Our S-SGD-5x & & & 23467  \\
Our S-SGD-10x & & & 46933  \\
Our S-SGD-20x & & & 93865  \\
\bottomrule
\end{tabular}
\label{tab:data_comparison}
\end{table*}

It is widely recognized that the performance of neural networks heavily relies on both the quantity and quality of the data employed in the training stage.
Despite efforts of the data collection within the field of glass segmentation, there currently lacks a sufficiently large-scale glass surface segmentation dataset due to the time-consuming and labor-intensive nature of image acquisition and annotation processes.
Three glass surface segmentation datasets are publicly available, including GDD \cite{mei2020don}, GSD \cite{lin2021rich}, and GSD-S \cite{lin2022exploiting}.  However, each of these datasets comprises no more than 5,000 images, thus imposing limitations on the segmentation accuracy of neural networks. We release the largest glass segmentation dataset to date, termed S-GSD, consisting of 168k images with precise masks. Table \ref{tab:data_comparison} presents a detailed comparison of various glass surface segmentation datasets.
\subsection{Data Construction}
\begin{figure*}[t]
  \centering
  \includegraphics[width=7.15in]{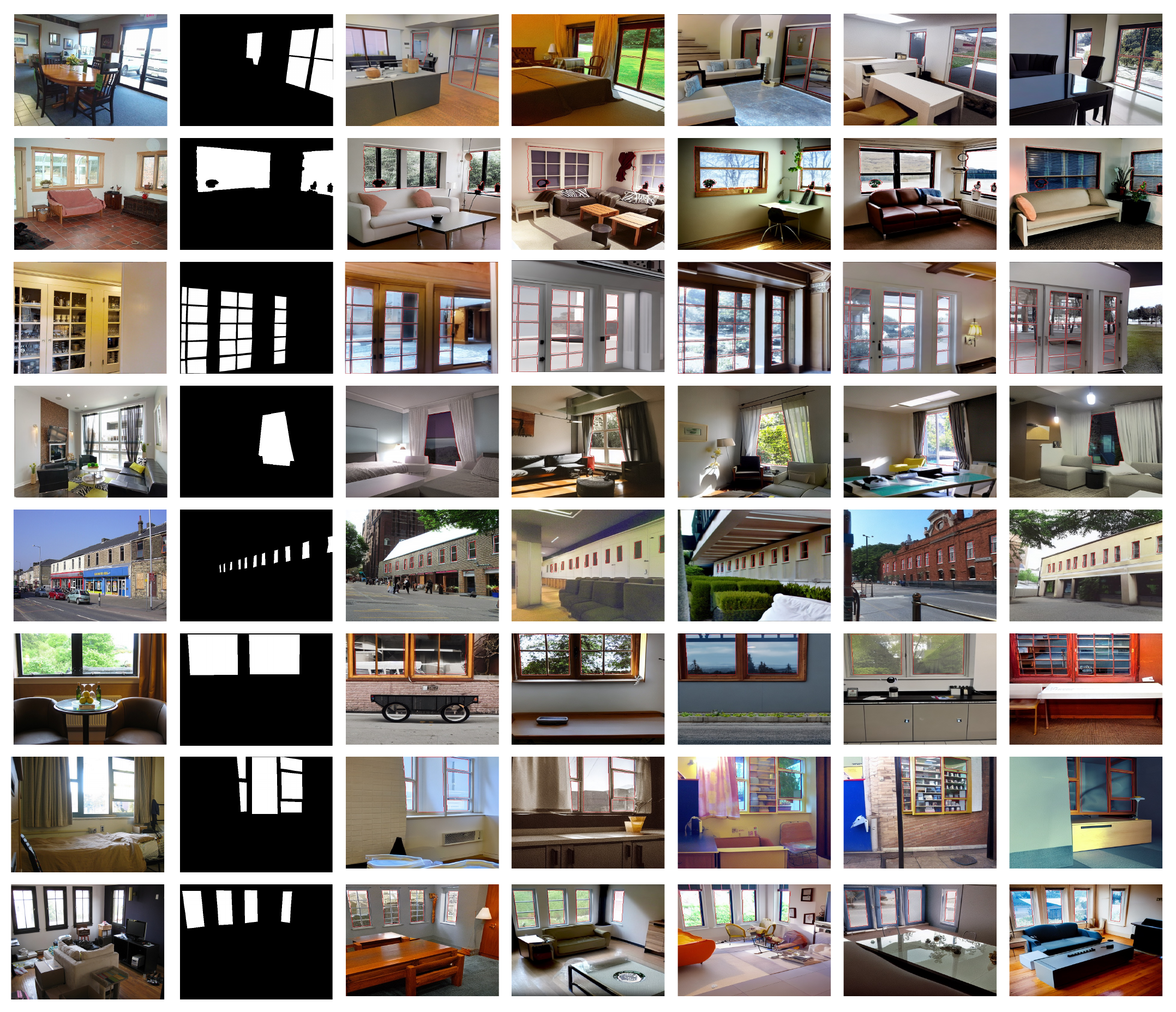}
  \caption{Visual examples of synthetic data. The first and second columns refer to the real data and corresponding mask, respectively. The rest of the columns are the synthetic data. We draw the \textcolor{red}{\textbf{Red Edge}} of the mask on the synthesized image to demonstrate the precise alignment between the glass region and the mask.}
  \label{fig:synthetic_data}
\end{figure*}

We exploit the VFM, Stable Diffusion Model, to automatically synthesize S-GSD dataset with high-quality mask annotation, entirely devoid of human intervention.
Specifically, we utilize the ControlNet with Stable Diffusion to generate massive high-quality images by using the mask annotations in the publicly available dataset as control conditions; thus, the generated image-mask pair can construct a new segmentation training sample. In order to achieve domain customization and match the distribution between the synthetic data and the real data, we finetune the ControlNet with real data while keeping the stable diffusion frozen. Simultaneously, we use plenty of language prompts (23 in total) to increase the diversity of synthesized images. Three examples of language prompts for image generation are as follows: 
“\textit{a photo of a clean} \textless\text{object}\textgreater”, “\textit{a close-up photo of the }\textless\text{object}\textgreater”, “\textit{a rendition of the }\textless\text{object}\textgreater”. Here, the “\textless\text{object}\textgreater” refers to \textit{transparent glasses}. Hence, the generation process requires the control inputs of the segmentation mask and the language prompt. Note that we don’t use the validation mask during the process of finetuning the ControlNet and synthetic data generation to avoid the leakage of validation data. Our synthetic dataset S-GSD consists of four scales, including 4k, 24k, 47k, and 94k image-mask paris, respectively.
Fig. \ref{fig:synthetic_data} shows some synthetic images in several mask control conditions. Given a specific mask condition, we can generate high-quality images where the position of the glasses is strictly follows the segmentation map layout while the background is full of diversity. 

\setlength{\belowcaptionskip}{-0.2cm}


\subsection{Data Analysis}
Surprisingly, the synthesized data exhibits three prominent advantages: \textbf{(1) The physical nature}. The generated glass possesses natural properties, including transparency, reflection, and refraction, akin to real glass, adhering to the principles of physical optics, which is displayed in Fig. \ref{fig:synthetic_data}. \textbf{(2) The precise mask}.The glass in the generated images is precisely aligned with the mask control condition, resulting in images that possess accurately annotated segmentation maps. We draw the red edge of the mask on the synthesized image to demonstrate the precise alignment between the glass region and the mask. \textbf{(3) The rich diversity}. The generated images encompass a diverse array of background contents, thus providing a more rich and varied visual representation for neural networks. Table 1 shows that the image complexity \cite{feng2022ic9600} in our S-GSD dataset is obviously higher than those from GDD, yet slightly lower than the other two datasets.



\section{Proposed Method}
\label{Methodology}
In this section, our primary goal is to make full use of the knowledge of foundation models to design a simple but accurate segmentor for glass surfaces. The model architecture is illustrated in Sec. \ref{sec: Model architecture}. We also propose a Discerning Query Selection module that can improve the accuracy of segmentation, which is introduced in Sec. \ref{discerning query selection}. Finally, the loss function is discussed in Sec. \ref{Loss Function}. 
\begin{figure}[t]
  \centering
  \includegraphics[width=3.35in]{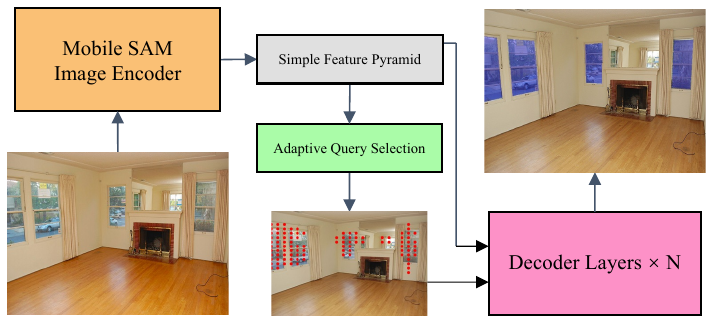}
  \caption{The architecture of proposed GEM. The discerning query selection is to predict the foreground and its corresponding features will be used to initialize the decoder’s query}
  \label{fig:framework}
\end{figure}

\subsection{Model architecture}
\label{sec: Model architecture}
\cite{han2023segment} had evaluated the performance of SAM in the glass detection field, and it proves the visual foundation model cannot directly segment glass without any modification.
Hence, we make a enhancement based on SAM and proposed the GEM network. GEM adopts a query-based encoder-decoder architecture that consists of an image encoder, a simple feature pyramid, discerning query selection, and a mask decoder, as shown in Fig. \ref{fig:framework}. 

In order to harness the capabilities acquired by large models from massive corpora, we employ the image encoder derived from MobileSAM for GEM-Tiny and SAM for GEM-Base.

To boost the performance for glass surface segmentation, we use only the last feature map from the image encoder to produce multi-scale feature maps via a simple feature pyramid following ViTDet. Specifically, we generate feature maps of scales 1/8, 1/4, and 1/32 using deconvolution of strides 2 and 4 and maxpooling of strides 2, respectively.

Given these hierarchical feature maps, we simply predict masks via a mask decoder used in MaskDINO \cite{li2023mask} with simplified improvement. The fusion operation for generating the pixel embedding map in MaskDINO is removed, and the feature map of scale 1/4 is directly appointed as the role of the pixel embedding map. We found this minimal modification did not harm the performance. Eventually, we obtain an output mask by dot-producting each content query embedding from the mask decoder with the pixel embedding map. 

To summarize, an image $\mathcal{I} \in \mathbb{R}^{H\times W\times3}$ is inputted to the image encoder, and we can obtain four-scale feature maps \textit{C2}, \textit{C3}, \textit{C4}, and \textit{C5} via a simple feature pyramid $\mathcal{P}$, of which the resolutions are 1/4, 1/8, 1/16, and 1/32 of the input image, respectively.
Then the mask decoder takes the queries $\mathcal{Q} \in \mathbb{R}^{N\times256}$ and the flattened three high-level feature maps \textit{C3}, \textit{C4}, and \textit{C5} as inputs and updates queries $\mathcal{Q}$.
Finally, the updated queries $\mathcal{Q}$ dot-product with the pixel embedding map \textit{C2} to obtain a predicted mask $\mathcal{M}$. The whole process can be formulated as follows:
\begin{equation} 
\textit{C2, C3, C4, C5} = \mathcal{P}( \mathcal{E}(\mathcal{I})),
\end{equation}
\begin{equation}
M = \textit{C2} \otimes \mathcal{D}(\mathcal{Q}, Flatten(\textit{C3,C4,C5})),
\end{equation}
where $\mathcal{E}$ is the image encoder and $\mathcal{D}$ is the mask decoder. The $\otimes$ indicates the dot production. Note that the prediction masks are output at each decoder layer.

\subsection{Discerning Query Selection}
\label{discerning query selection}
\begin{figure}[t]
  \centering
  \includegraphics[width=3.35in]{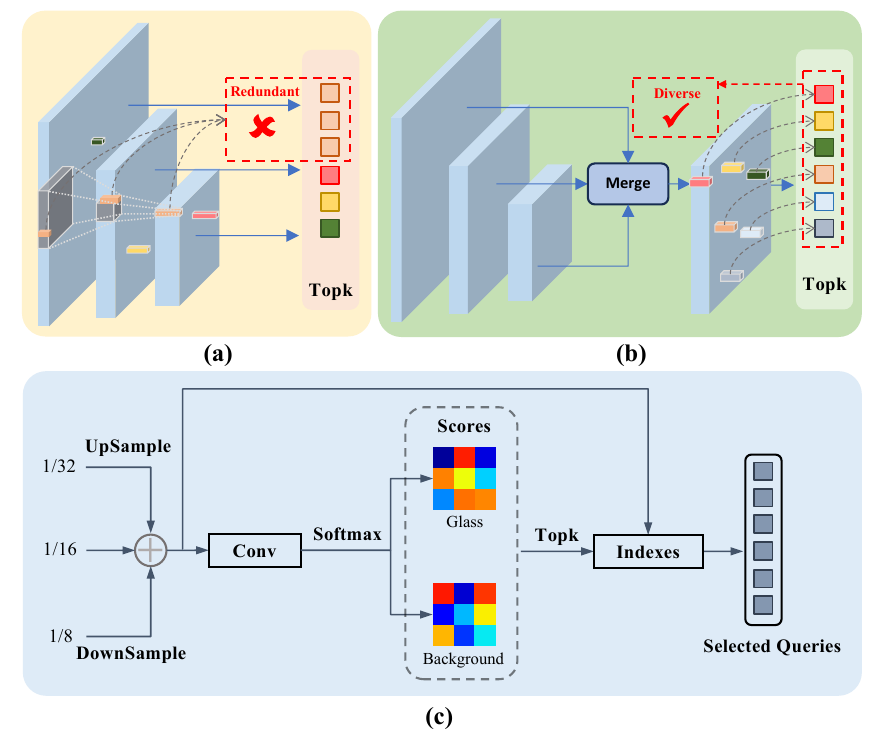}
  \caption{Illustration of the discerning query selection module. (a) MaskDINO's query selection method chooses top-k queries based on separate intermediate features, potentially resulting in redundant information when different queries are located in the same receptive field. (b) Our approach involves merging different-level features before selecting queries on the integrated feature, minimizing redundancy. (c) Detailed structure of our discerning query selection module.}
  \label{fig:DQS}
\end{figure}

The query initialization has been proven vital for the final prediction results \cite{zhang2022dino, li2023mask}. To achieve accurate glass segmentation, we employ a discerning query selection module that is based on confidence-aware query initialization for the decoder, as shown in Fig.~\ref{fig:DQS} (c). Different from MaskDNIO which selects queries from separate intermediate features, as shown in Fig.~\ref{fig:DQS} (a), we first merge them and operate on the integrated features, as shown in Fig.~\ref{fig:DQS} (b). The query selection method of MaskDINO has the potential drawback that the selected queries from different levels are located in the same receptive field, which may introduce redundant information. This redundancy is particularly problematic in glass segmentation, where the foreground and background are frequently very similar due to the lack of distinctive texture and color information in glass. Redundant information will reduce the differentiation between glass and other regions among the selected queries. To verify this hypothesis, we train MaskDINO using various query selection source strategies on our S-GSD dataset, as presented in Table~\ref{tab:ablation_maskdino}. 
\begin{table}[b]
\centering
\tabcolsep=0.25cm
\caption{Segmentation performance on the S-GSD dataset using different query selection source for MaskDINO. }
\begin{tabular}{lccccc}
\toprule
Source of Feature Selection   & IoU $\uparrow$    & $F_{\beta} \uparrow$ & MAE $\downarrow$  & BER $\downarrow$   \\
\midrule
C3+C4+C5             & 0.654 & 0.792 & 0.061 & 13.63     \\
C3                   & 0.657 & 0.792 & 0.061 & 13.32     \\
C4                   & 0.659 & 0.800 & 0.059 & 13.38     \\
\bottomrule
\end{tabular}
\label{tab:ablation_maskdino}
\end{table}
Our findings indicate that the original MaskDINO, which utilizes C3, C4, and C5 features to obtain queries, yields the poorest results. In contrast, utilizing only C3 or C4 features results in better performance. We believe that using a single-level feature can encompass more regions, allowing the queries to include more foreground and background information simultaneously.
Hence, we aggregate the \textit{C3} and \textit{C5} layers with \textit{C4} layer by downsample and upsample operations, leading to the aggregated feature $\mathbf{f}$.
After that, we obtain the feature-wise classification result $\mathbf{P}^{2 \times h \times w}$ based on the $\mathbf{f}$ through the $Softmax$ operation. The $\mathbf{P}$ provides a coarse confidence prediction for glass and background regions and is optimized under the supervision of ground truth.

To obtain high-quality queries, we rank all confidence scores $\mathbf{S} \in \mathbb{R}^{2hw}$ and select the features corresponding to the top-k scores as our queries. Our discerning query selection is designed to empower the decoder by leveraging pre-classification prior content features. This selection process is effective in diverse scenarios, allowing us to adaptively identify features that significantly contribute to classification, thereby assigning them as initialized queries. This becomes particularly crucial when dealing with challenging scenarios where the color and texture of the glass are similar to those of the background, posing difficulties in discrimination. Further analysis of its functionality will be presented in Section~\ref{visualization for query selection}.

\subsection{Loss Function}
\label{Loss Function}
During the training phase, we adopt a loss function following Mask DINO \cite{li2023mask} and incorporate an additional classification loss $\mathcal{L}_{q}$ tailored for query selection. The overall loss is structured as a weighted sum of three primary loss components: $\lambda_{cls}\mathcal{L}_{cls} + \lambda_{L1}\mathcal{L}_{L1} + \lambda_{giou}\mathcal{L}_{giou} + \lambda_{ce}\mathcal{L}_{ce} + \lambda_{dice}\mathcal{L}_{dice} + \lambda_{q}\mathcal{L}_{q}$. Specifically, the L1 loss $\mathcal{L}_{L1}$ and GIOU loss $\mathcal{L}_{giou}$ are employed for box regression, while $\mathcal{L}_{cls}$ represents a focal loss designed for classification purposes. The mask prediction aspect utilizes both cross-entropy $\mathcal{L}_{ce}$ and dice loss $\mathcal{L}_{dice}$. Additionally, $\mathcal{L}_{q}$ is defined as a cross-entropy loss calculated between $\mathbf{P}$ and the ground truth. The Specific hyper-parameters setting will be provied in Section~\ref{sec:4.1}.

\section{Experiments and Discussion}
\label{Experiments and Discussion}

\subsection{Experimental Setup}
\label{sec:4.1}
\noindent
\textbf{Datasets.}
We design and conduct extensive experiments to verify the performance of GEM and the effectiveness of the S-GSD dataset. We select the newest and large-scale glass surface segmentation dataset, GSD-S \cite{lin2022exploiting}, to evaluate our GEM’s performance. This dataset consists of 3911 training samples and 608 test samples. Besides, we compare our proposed S-GSD dataset with two real glass datasets, GSD and GDD, via transfer learning, which demonstrates that our auto-generated glass dataset can express similar improvements compared with real glass datasets annotated manually.

\noindent
\textbf{Implementation Details.} For fair comparison, we keep the evaluation metric and image pre-processing the same with GlassSemNet \cite{lin2022exploiting}. 
The input data is first uniformly resized to the size of 384 × 384 before applying normalization, except where noted. 
Following \cite{han2023internal}, we don't use conditional random fields (CRF) as a post-processing. The models are trained 80 epochs in total on 4 NVIDIA GeForce RTX 3060 GPUs. We utilize the ViT-Tiny in MobileSAM \cite{zhang2023faster} and the ViT-Base in SAM \cite{kirillov2023segment} as our backbone's pre-trained models. The detailed training parameters in terms of GEM-Tiny and GEM-Base are provided in Table \ref{tab:training_parameters}.

\begin{table}[t]
\centering
\tabcolsep=0.65cm
\caption{The training parameters for GEM.}
\begin{tabular}{l|c|ccc}
\toprule
Training Parameters & GEM-Tiny & GEM-Base   \\
\midrule
learning rate  & 2e-4 & 5e-5  \\
weight decay     & 0.05 & 0.05 \\
batch size   & 32 & 32   \\
image size & 384 & 384 \\
epochs for pretrain & 160 & 160  \\
epochs for finetune & 80 & 80  \\
\# queries & 100 & 100  \\
dropout & 0.0 & 0.0 \\
$\lambda_{cls}$ & 4 & 4 \\
$\lambda_{L1}$ & 5 &5 \\
$\lambda_{giou}$ & 2&2 \\
$\lambda_{ce}$ & 5&5 \\
$\lambda_{dice}$ & 5&5 \\
$\lambda_{q}$ & 80&80 \\
\bottomrule
\end{tabular}
\label{tab:training_parameters}
\end{table}

\subsection{Comparisons with the State-of-the-Arts}
\label{sec:4.2}
We present a comprehensive evaluation of GEM against a diverse set of state-of-the-art semantic segmentation methods, encompassing mainstream transformer-based models like SETR \cite{zheng2021rethinking}, Segmenter \cite{strudel2021segmenter}, Swin \cite{liu2021swin}, SegFormer \cite{xie2021segformer}, Mask2Former \cite{cheng2021mask2former}, Mask DINO \cite{li2023mask}, and FASeg \cite{he2023dynamic}, as well as specialized models for glass segmentation including GDNet \cite{mei2020don}, GlassNet \cite{lin2021rich}, and GlassSemNet \cite{lin2022exploiting}.
Most methods are re-trained on GSD-S, following the default training settings specified in the original papers.
The results presented in Table \ref{tab:GSD-S-Comparisons} highlight the effectiveness of GEM across four evaluation metrics. GEM outperforms all competing models, showcasing its superiority in glass surface segmentation. Notably, when compared to the GlassSemNet, which was the previous state-of-the-art method on the GSD-S dataset, both GEM-Tiny and GEM-Base achieve substantial improvements, with a notable 1.7\% and 2.1\% enhancement in the metric IoU, respectively, along with improvements across other metrics. 
Furthermore, GEM-Tiny achieves remarkably high accuracy with minimal parameters and FLOPs, achieving an IoU of 74.7\%. After pretraining with our S-GSD dataset, the IoU further improves by 2.3\%.
Besides, we conduct comparisons of the FPS (Frame Per Second) across different methods. GEM-Tiny and GEM-Base exhibited processing speeds three times and two times faster, respectively, than GlassSemNet, validating the efficiency of our model. 

\begin{figure*}[]
  \centering
  \includegraphics[width=6.6in]{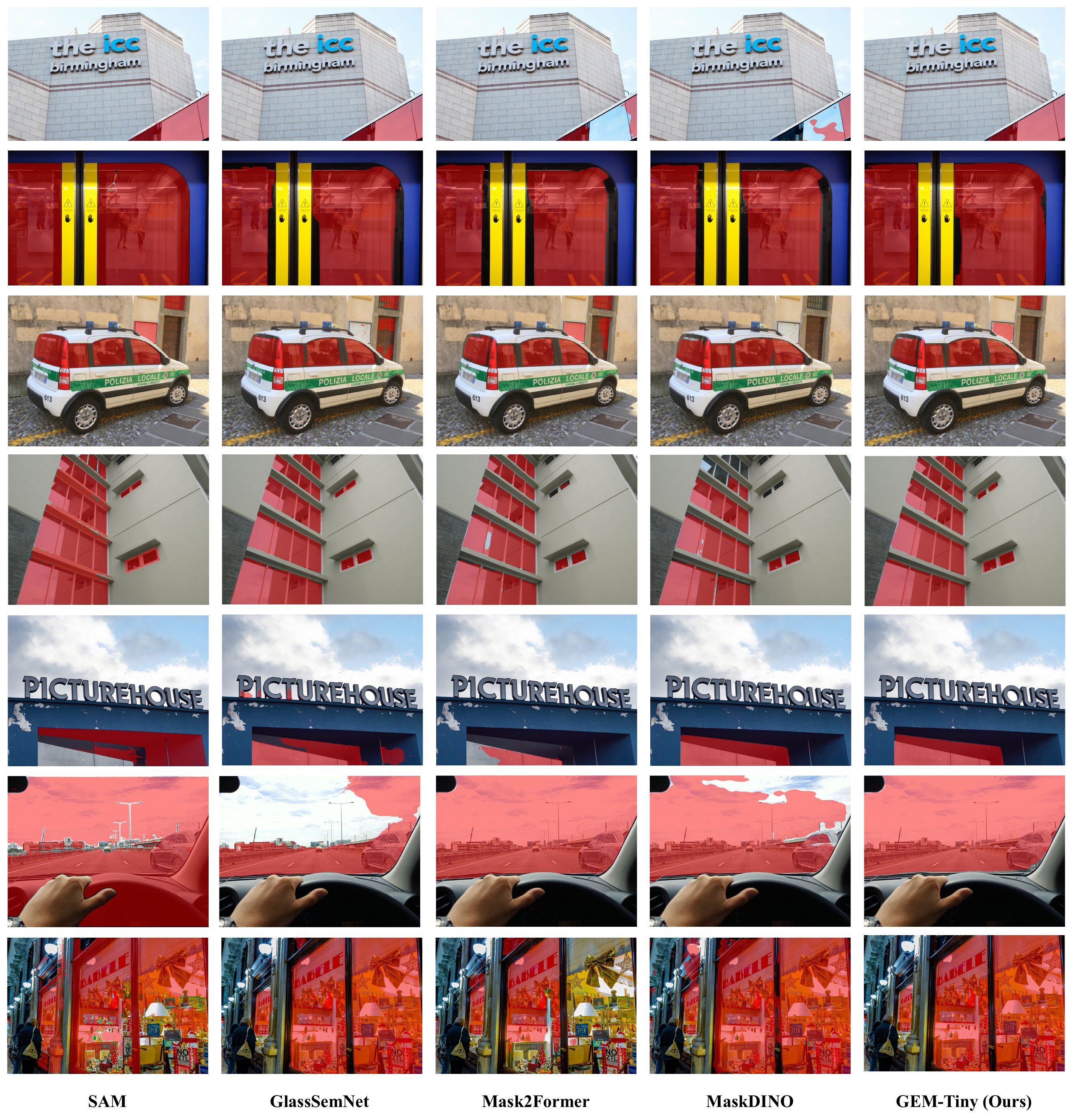}
  \caption{Visualization comparisons of our GEM-Tiny and other state-of-the-art methods tested on real-world scenes.}
  \label{fig:realdata}
\end{figure*}

\begin{figure}[t]
  \centering
  \includegraphics[width=3.5in]{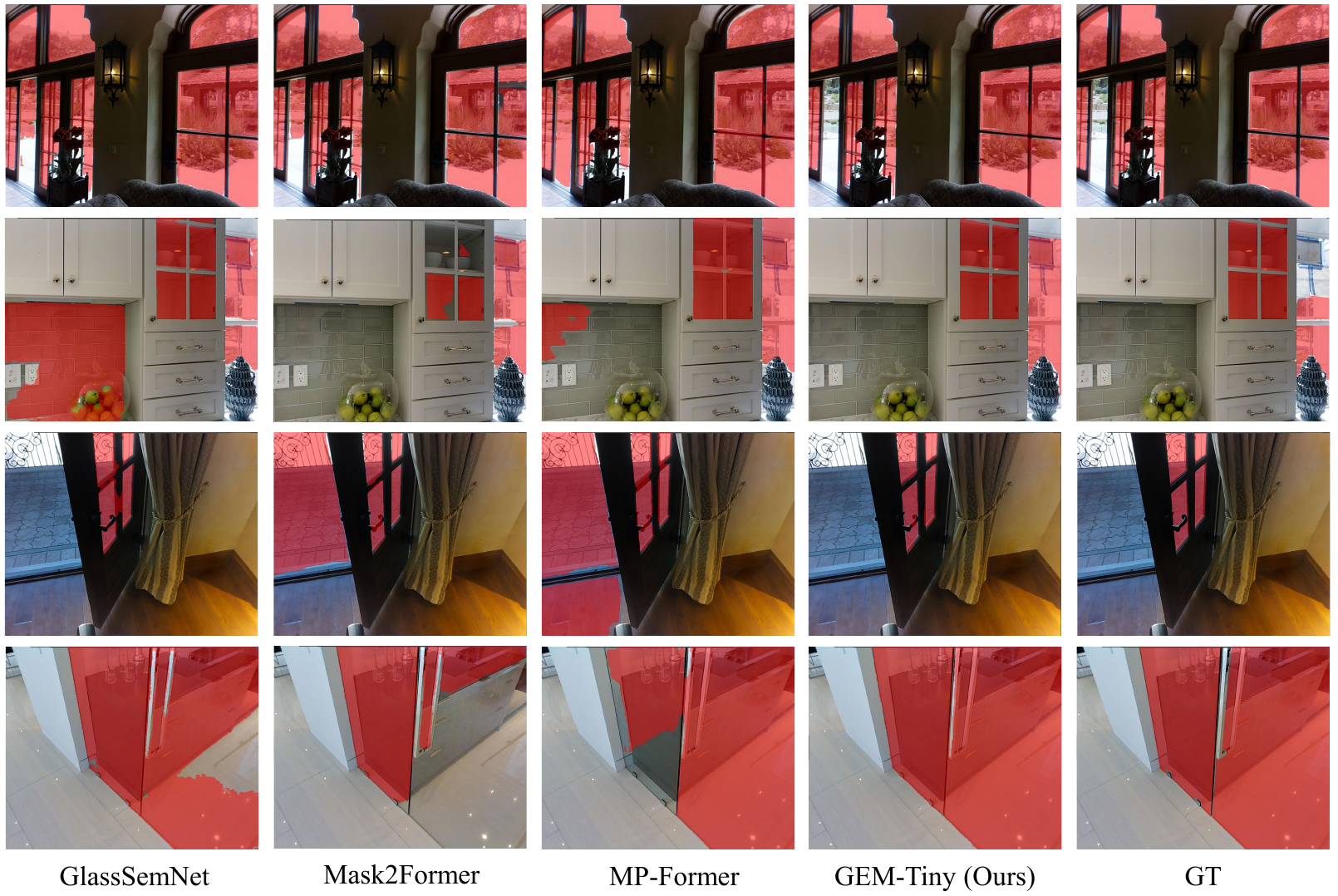}
  \caption{Qualitative comparisons of our GEM with other methods on GSD-S validation set.}
  \label{fig:qualitative_comparison}
\end{figure}

In Fig.~\ref{fig:qualitative_comparison}, we present qualitative comparisons between our proposed method and other state-of-the-art networks. We highlight segmentation results in four distinct scenarios, showing the effectiveness of our GEM-Tiny alongside three other methods. Notably, our approach excels in distinguishing glass regions, outperforming other methods that often struggle, particularly when faced with similarities in color and texture features between glasses and the background. 
We also conducted additional tests on real-world scenes, which proves the generalization of our model. The test images are derived from the SA-1B dataset\footnote{https://segment-anything.com/dataset/index.html}, which is released by SAM. This dataset is composed of images licensed from a photo provider that works directly with photographers.
The segmentation results of SAM, GlassSemNet, Mask2Former, MaskDINO, and our GEM-Tiny are presented in Fig.~\ref{fig:realdata}. For SAM, we employ Grounded-SAM\footnote{https://github.com/IDEA-Research/Grounded-Segment-Anything}, utilizing text prompt as the "window" for targeted segmentation. While SAM achieves the best segmentation results in the first four scenes, its performance deteriorates to varying degrees in the last six scenarios. We attribute this decline to SAM's class-agnostic nature. Our GEM-Tiny outperforms the other segmentation methods on these scenes, demonstrating accurate region judgment and superior edge delineation detail. Among the other three segmentation methods, GlassSemNet, Mask2Former, and MaskDINO, each exhibits shortcomings, experiencing failures in specific scenes and often producing incorrect or missing segmentation.

\begin{table*}[]
\centering
\tabcolsep=0.45cm
\caption{The evaluation results on GSD-S. Some scores of competing methods are taken from GlassSemNet. ``pret'' indicates pretraining on our proposed S-GSD dataset.}
\begin{tabular}{>{}p{3cm} >{\centering\arraybackslash}p{1.5cm} >{\centering\arraybackslash}p{1.5cm} >{\centering\arraybackslash}p{1.0cm} >{\centering\arraybackslash}p{1.0cm} >{\centering\arraybackslash}p{1.0cm} >{\centering\arraybackslash}p{1.0cm}>{\centering\arraybackslash}p{1.0cm}}
\toprule
Methods    & Params(M)$\downarrow$ & FLOPs(G)$\downarrow$ & IoU$\uparrow$ & $F_{\beta}\uparrow$ &MAE$\downarrow$  &BER$\downarrow$  &FPS$\uparrow$ \\
\midrule
SCA-SOD \cite{siris2021scene}      & - & -   & 0.558  & 0.689     & 0.087 & 15.03  & -\\
SETR \cite{zheng2021rethinking}      & - & -       & 0.567  & 0.679     & 0.086 & 13.25  & -\\
Segmenter \cite{strudel2021segmenter}  & - & -      & 0.536  & 0.645     & 0.101 & 14.02  & -\\
Swin \cite{liu2021swin}        & - & -       & 0.596  & 0.702     & 0.082 & 11.34  & -\\
ViT \cite{yuan2021tokens}     & - & -           & 0.562  & 0.693     & 0.087 & 14.72  & -\\
Twins \cite{chu2021twins}      & - & -  & 0.590  & 0.703     & 0.084 & 12.43  & -\\
SegFormer \cite{xie2021segformer}  & - & -   & 0.547  & 0.683     & 0.094 & 15.15  & -\\
MaskFormer \cite{cheng2021maskformer}    & 41.3 & 31.6     & 0.707  & 0.826    & 0.043 & 10.91 & 28.90 \\
Mask2Former\cite{cheng2021mask2former}    & 44.0 & 38.4     & 0.732  & 0.838     & 0.043 & 8.93 & 16.20 \\
Mask DINO \cite{li2023mask}   & 44.0 & 36.6  & 0.687   &  0.816     & 0.049  & 11.67 & 13.01  \\
FASeg \cite{he2023dynamic}     & 50.5 & 40.2  & 0.725  & 0.843     & 0.048 & 10.26 & 11.32 \\ 
MP-Former \cite{zhang2023mp}       & 43.9 & 38.4    & 0.734  & 0.827     & 0.042 & 8.67 & 16.59 \\
NAT \cite{hassani2023neighborhood}       & &         & 0.730  & 0.846     & 0.041 & 10.16 & 16.68 \\
\midrule
GDNet \cite{mei2020don}        & - & -   & 0.529  & 0.642     & 0.101 & 18.17 & -\\
GlassNet \cite{lin2021rich}     & - & -    & 0.721  & 0.821     & 0.061 & 10.02  & -\\
GlassSemNet \cite{lin2022exploiting}   &36.1 & 141.2 & 0.753  & 0.860     & 0.035 & 9.26  & 5.64 \\
\midrule
GEM-Tiny w/o pret      & 21.6 & 11.5 & 0.747     & 0.845  & 0.039  & 8.52 & 16.09 \\
GEM-Tiny w/ pret      & 21.6 & 11.5  & 0.770 & 0.865 & 0.032 & \textbf{8.21} & 16.09 \\
GEM-Base w/o pret     & 102.7 & 63.7    & 0.756 & {0.862} & {0.034} & 9.31 & 11.55 \\
GEM-Base w/ pret     & 102.7 & 63.7    & \textbf{0.774} & \textbf{0.865} & \textbf{0.029} & 8.35 & 11.55 \\
\bottomrule
\end{tabular}
\label{tab:GSD-S-Comparisons}
\end{table*}

\begin{table}[]
\centering
\tabcolsep=0.29cm
\caption{The results of zero-shot \& finetune on GSD-S validation set compared with three different pretraining datasets.}
\begin{tabular}{lcccccc}
\toprule
Paradigm & Dataset  & IoU $\uparrow$    & $F_{\beta} \uparrow$ & MAE $\downarrow$  & BER $\downarrow$   \\
\midrule
             &  GDD       & 0.481 & 0.620 & 0.179 & 20.42     \\
Zero-Shot    &  GSD       & \textbf{0.714} & 0.811 & \textbf{0.054} & \textbf{10.30} \\
             &  S-GSD-1x  & 0.703 & \textbf{0.819} & 0.215 & 10.79    \\
\midrule
             &  GDD       & 0.743 & 0.843 & 0.039 &  8.55    \\
Finetune     &  GSD       & 0.753 & 0.851 & \textbf{0.037} & 8.72     \\
             &  S-GSD-1x  & \textbf{0.755} & \textbf{0.852} & 0.038 & \textbf{8.39}    \\
\bottomrule
\end{tabular}
\label{tab:Effeciveness_of_S-GSD}
\end{table}

\subsection{Effectiveness of auto-generated S-GSD}
\label{sec:4.3}
It is evident that large-scale datasets can boost the performance of detection and segmentation tasks via transfer learning \cite{hao2023language}. 
To verify the quality and effectiveness of the generated S-GSD, we compared it with two real high-quality public-available datasets, GDD and GSD, in terms of the performance of zero-shot and finetune on GSD-S with the GEM-Tiny model. The number of images in GDD, GSD, and S-GSD-1x is 3916, 4102, and 3912, respectively, so the experimental results are fairly comparable. 

As for the zero-shot paradigm, the performance of GSD and S-GSD-1x exhibits an obvious increase compared with the GDD, while the metrics of GSD and S-GSD-1x are quite similar. Regarding the finetuning on GSD-S, our auto-generated pretraining dataset S-GSD-1x outperforms other real public datasets in three metrics, including IoU, $F_{\beta}$, and BER. Additionally, the metric MAE on S-GSD-1x and GSD are numerically close (0.038 vs. 0.037). These results prove that our S-GSD-1x dataset is as photorealistic as the real data, and it is indeed suitable to be used as the pretraining data. 

Moreover, to further validate the effectiveness of S-GSD as pretraining data, we pretrain existing methods on S-GSD-1x and then finetune them on GSD-S, which is shown in Table \ref{tab:pretrain_on_existing_method}. Results show improvements for all methods, but they still perform worse than GEM-T. 

\begin{table}[h]
\centering
\renewcommand{\arraystretch}{0.8}
\tabcolsep=0.1cm
\caption{The effectiveness of S-GSD as pretraining data on existing methods.}
\begin{tabular}{>{\small}p{2.2cm} >{\centering\arraybackslash}p{3cm} > {\centering\arraybackslash}p{3cm}}
\toprule
Methods   & IoU $\uparrow$ (w/o Pretrain)& IoU $\uparrow$ (w Pretrain)  \\
\midrule
MaskDINO & 0.687   & 0.705   \\
MaskFormer  & 0.707   & 0.723  \\
Mask2Former  & 0.732   & 0.742   \\
\midrule
GEM-T (Ours)  & \textbf{0.744}  & \textbf{0.755}  \\
\bottomrule
\label{tab:pretrain_on_existing_method}
\end{tabular}
\end{table}

\subsection{The impact of synthetic data scale}
\label{The impact of synthetic data scale}
With the increase in pretraining data scale, pretrained models acquire more comprehensive visual semantic representations, leading to enhanced performance in downstream tasks after fine-tuning \cite{hao2023language}. We explore the impact of synthetic data scale on glass surface segmentation. Specifically, we generate four pretraining datasets that are 1x, 5x, 10x, and 20x larger than the GSD-S, respectively. These pretraining datasets are used to pretrain and finetune our GEM model. We pretrain 160 epochs in the pretraining stage and 80 epochs in the finetuning stage. Concerning the zero-shot for GEM-Tiny, the IoU and $F_{\beta}$ increase by 2.3\% and 2.5\% when the data scale expands from 1x to 20x, respectively. The same trend can be observed in the performance of finetune for GEM-Tiny, which increased by 1.5\% and 1.3\%, respectively. As for the GEM-Base, it is also clear that the performance of zero-shot and finetune is steadily improving when the scale of pretraining data increases continuously. However, when the data scale expands from 10x to 20x, the improvement in finetuning for the GEM-Base is almost saturated, which could be a bottleneck to improvement in transfer learning. Overall, as the scale of synthetic data grows, both zero-shot and fine-tuning performances demonstrate a consistent and significant performance improvement.

\begin{table}[]
\centering
\tabcolsep=0.23cm
\caption{The ablation study of the scale of the synthetic data compared with GSD-S.}
\begin{tabular}{llcccccc}
\toprule
Model & Paradigm & Scale  & IoU $\uparrow$    & $F_{\beta} \uparrow$ & MAE $\downarrow$  & BER $\downarrow$   \\
\midrule
&             &  1 $\times$   & 0.703 & 0.819 & 0.215 & 10.79       \\
GEM-T &Zero-Shot    &  5 $\times$      & 0.712 & 0.828 & 0.211 & 11.22         \\
&             &  10 $\times$     & 0.723 & 0.826 & 0.211 & \textbf{10.07}     \\
&             &  20 $\times$     & \textbf{0.726} & \textbf{0.844} & \textbf{0.206} & 10.98     \\
\midrule
&             &  1 $\times$   & 0.755 & 0.852 & 0.038 & 8.39       \\
GEM-T &Finetune    &  5 $\times$      & 0.757 & 0.855 & 0.035 & 8.54         \\
&             &  10 $\times$     & 0.764 & \textbf{0.866} & 0.034 & 8.62        \\
&             &  20 $\times$     & \textbf{0.770} & 0.865 & \textbf{0.032} & \textbf{8.21}        \\
\midrule
&             &  1 $\times$   & 0.701 & 0.808 & 0.218 & 11.08       \\
GEM-B &Zero-Shot    &  5 $\times$      & 0.720 & 0.827 & 0.210 & 10.84        \\
&             &  10 $\times$     & 0.725 & 0.830 & 0.209 & 11.20     \\
&             &  20 $\times$     & \textbf{0.729} & \textbf{0.839} & \textbf{0.206} & \textbf{10.73}     \\
\midrule
&             &  1 $\times$   & 0.766 & 0.873 & 0.031 & 9.44       \\
GEM-B &Finetune    &  5 $\times$      & 0.769 & 0.858 & 0.032 & 8.16         \\
&             &  10 $\times$     & \textbf{0.774} & \textbf{0.868} & 0.032 & 8.56        \\
&             &  20 $\times$     & \textbf{0.774} & 0.865 & \textbf{0.029} & \textbf{8.35}       \\
\bottomrule 
\end{tabular}
\label{tab:transfer111}
\end{table}

\subsection{Ablation Study}
\textbf{Ablation on DQS module.}
The DQS module has influenced the segmentation results in two aspects. One is that it adds an additional binary classification loss from multi-scale features, and the other is that it selects high-quality content features for query initialization in the decoder. To verify the effectiveness of the DQS, we conduct some experiments as shown in Table \ref{tab:ablation} (a). The metric IoU and $F_{\beta}$ decrease by 1.0\% and 1.8\% when only removing the DQS initialization part, respectively, which proves the improvement of DQS initialization. Surprisingly, we saw a slight increase when removing the extra binary classification loss. We conjecture that the additional loss tends to make the distribution of image features closer to the ground-truth mask, which increases the difficulty of refining randomly initialized queries in the decoder.

\noindent
\textbf{Ablation on query selection source for DQS.}
Our DQS first integrate three level features and choose queries on the merged feature. To explore other potential source strategies, we conduct experiments as shown in Table \ref{tab:ablation} (b). The C3+C4+C5 strategy, which selects top-k queries while considering all three feature levels simultaneously (same to MaskDINO), achieves the worst performance. This result aligns with our analysis in Section~\ref{discerning query selection}. Using only C3 or C4 individually also yields inferior results compared to our approach. This is because our strategy effectively fuses multi-level information, and also ensuring the diversity of the selected queries.

\noindent
\textbf{Language prompt in data generation.}
In the data generation process, the language prompt is required as the control inputs, and the multiple prompts can increase the data diversity. We explore the influence of language prompts on the zero-shot and fine-tuning outcomes. In Table \ref{tab:ablation} (c), when utilizing multiple prompts as the language control input, four metrics are consistently superior to the single prompt in the zero-shot setting. As for the fine-tuning stage, the IoU, MAE, and BER metrics of using multiple prompts are better than using a single prompt, and the $F_{\beta}$ metric of these two settings is relatively close. 

\noindent
\textbf{Ablation on the SAM pre-trained model.}
To fully exploit the knowledge of large foundation models, we select the image encoders of MobileSAM and SAM as our backbone. As shown in Table \ref{tab:ablation} (d), the MobileSAM can obtain
non-trivial improvement in terms of four metrics over the ViT-Tiny with ImageNet pre-trained model, e.g., 5.2\% IoU and 3.9\% $F_{\beta}$ increasement. At the same time, the performance of the ViT-Tiny with ImageNet pre-trained model is obviously higher than the one with random initialization on four metrics, which verifies the importance of image backbone initialization.

\begin{table}[]
\centering
\tabcolsep=0.25cm 
\caption{Ablation studies on the GSD-S validation set. }
\begin{tabular}{ccccccccc}
\toprule
\rowcolor{color0}
  &     & IoU $\uparrow$    & $F_{\beta} \uparrow$ & MAE $\downarrow$  & BER $\downarrow$   \\
\hline
\rowcolor{color1}
\multicolumn{6}{l}{(a) Ablation comparison of the proposed DQS.} \\
\hline
\rowcolor{color2}
Extra loss & DQS &&&& \\
\hline
\rowcolor{color2}
 \ding{55} & \ding{55}  & 0.739 & 0.840  & 0.041     & 8.73          \\
 \rowcolor{color2}
\checkmark   & \ding{55}   & 0.734  & 0.831  & 0.043 &  8.80       \\
\rowcolor{color2}
\checkmark            & \checkmark  & \textbf{0.747}     & \textbf{0.845}  & \textbf{0.039}  & \textbf{8.52}     \\

\hline
\rowcolor{color5}
\multicolumn{6}{l}{(b) Ablation comparison of Query Initialization in DQS.}  \\
\hline
\rowcolor{color6}
\multicolumn{2}{c}{Source of Feature Selection}&&&& \\
\hline
\rowcolor{color6}
\multicolumn{2}{c}{C3 + C4 + C5}  & 0.735 & 0.842 & 0.041 & 9.42       \\
\rowcolor{color6}
\multicolumn{2}{c}{C3} & 0.740 & 0.841 & 0.041 & 8.90      \\
\rowcolor{color6}
\multicolumn{2}{c}{C4} & 0.744     & 0.841  & 0.042  & \textbf{8.43}        \\
\rowcolor{color6}
\multicolumn{2}{c}{DQS} & \textbf{0.747}     & \textbf{0.845}  & \textbf{0.039}  & 8.52        \\

\hline
\rowcolor{color3}
\multicolumn{6}{l}{(c) Ablation of the language prompt used in the ControlNet.} \\
\hline
\rowcolor{color4}
Paradigm & Prompt  &&&&\\
\hline
\rowcolor{color4}
Zero-Shot &  Single   & 0.690 & 0.803 & 0.217 & 11.72       \\
\rowcolor{color4}
Zero-Shot   & Multiple      & 0.703 & 0.819 & 0.215 & 10.79         \\

\rowcolor{color4}
Finetune        &  Single   & 0.752 & \textbf{0.854} & 0.039 & 8.72       \\
\rowcolor{color4}
Finetune    &  Multiple     & \textbf{0.755} & 0.852 & \textbf{0.038} & \textbf{8.39}         \\
    
\hline
\rowcolor{color5}
\multicolumn{6}{l}{(d) Comparison of the SAM pre-trained model.}  \\
\hline
\rowcolor{color6}
\multicolumn{2}{c}{Pre-trained model}&&&& \\
\hline
\rowcolor{color6}
\multicolumn{2}{c}{Random}  & 0.551 & 0.681 & 0.091 & 16.90       \\
\rowcolor{color6}
\multicolumn{2}{c}{ImageNet} & 0.695 & 0.806 & 0.048 & 10.97      \\
\rowcolor{color6}
\multicolumn{2}{c}{MobileSAM} & \textbf{0.747}     & \textbf{0.845}  & \textbf{0.039}  & \textbf{8.52}        \\
\bottomrule
\end{tabular}
\label{tab:ablation}
\end{table}


\subsection{Visualization for Query Selection}
\label{visualization for query selection}
\begin{figure}[t]
  \centering
  \includegraphics[width=3.5in]{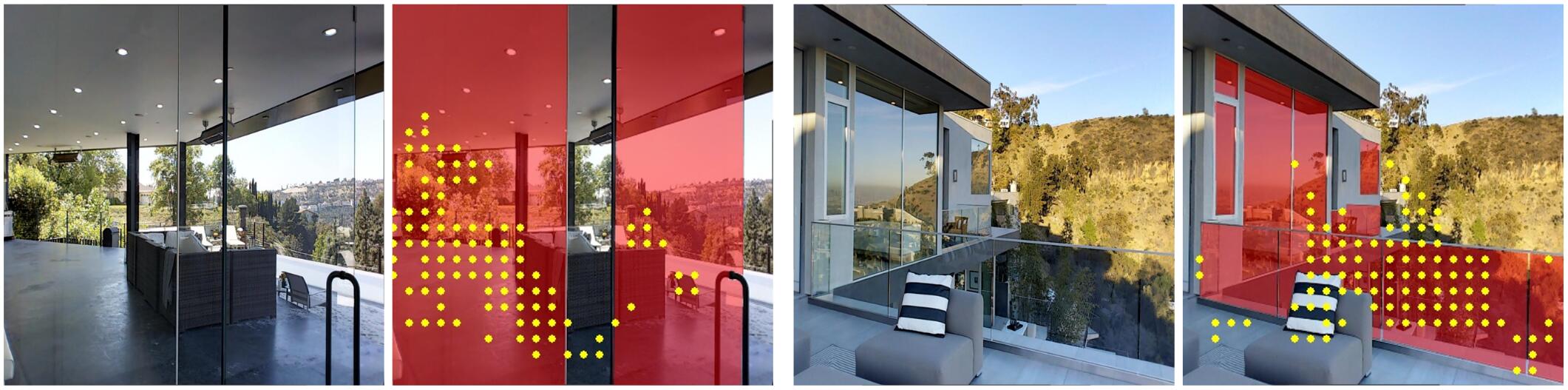}
  \caption{Visualization for discerning query selection. Given two scenarios, the left image depicts the original scene, while the right image highlights the glass mask in red, while the yellow dots indicate selected pixel positions}
  \label{fig:pointvis}
\end{figure}

To validate the effectiveness of our discerning query selection method, we choose two challenging scenarios, as shown in Fig.~\ref{fig:pointvis}. The example aligns with the description provided in Section \ref{discerning query selection}, where the region outside the glass and the glass area collectively constitute the same scene. This configuration poses a significant challenge for the model in distinguishing between the glass and background. Nevertheless, as depicted in Fig.~\ref{fig:pointvis}, the selected queries are predominantly situated within the glass region, resulting in a highly effective initialization of features. Though human observers may hesitate to discern the presence of glass on the left scene, given its extreme transparency and identical content to the background. However, our discerning query selection method successfully designates the left region as the primary selection target, 
which proves the designed query selection method can help our GEM better identify glass regions.

\section{Conclusion}
\label{Conclusion}
In this work, we are the first to propose exploring to the solution of glass surface segmentation by fully harnessing the capabilities of existing VFMs. By leveraging the prior knowledge of Stable Diffusion and the Segment Anything Model, we successfully address the challenges of high costs associated with data acquisition and annotation, along with the increasing complexity of model architecture within conventional deep learning approaches. We introduce a large-scale glass segmentation dataset, S-GSD, consisting of 168k images with precise masks. Additionally, we devise a simple glass surface segmentation framework, termed GEM, by transferring the precise segmentation capability of SAM.
Our GEM achieves high-precision glass segmentation on RGB images without extra cues like depth, polarized light, or thermal images.
Extensive experiments show that our GEM surpasses the previous state-of-the-art methods by a large margin. Also, our S-GSD exhibits substantial improvement on zero-shot learning and transfer learning settings. Hopefully, this completely new solution could bring inspiration to the study of visual perception integrated with AI-generated content.


 

\bibliographystyle{IEEEtran}
\bibliography{IEEEabrv,egbib}

\vfill

\end{document}